\pdfoutput=1
\documentclass{article}
\newcommand{\paperdir}{}
\IfFileExists{arxiv/main.tex}{\renewcommand{\paperdir}{arxiv/}}{}
\makeatletter
\edef\input@path{{\paperdir}}
\makeatother
\usepackage{iclr2027_conference,times}
\usepackage{amsmath,amssymb}
\usepackage{microtype}
\usepackage{graphicx}
\usepackage{placeins}
\usepackage{xcolor}
\usepackage{colortbl}
\usepackage{booktabs,tabularx}
\usepackage{enumitem}
\usepackage{algorithm}
\usepackage{algpseudocode}
\usepackage{tikz}
\usetikzlibrary{arrows.meta,positioning,calc,fit,backgrounds}
\PassOptionsToPackage{hyphens}{url}
\usepackage{hyperref}
\usepackage{url}

\definecolor{ink}{HTML}{233440}
\definecolor{selectblue}{HTML}{246E9B}
\definecolor{selectlight}{HTML}{EAF3F9}
\definecolor{composeorange}{HTML}{B96022}
\definecolor{composelight}{HTML}{FBF0E7}
\definecolor{quietgray}{HTML}{647582}
\definecolor{linegray}{HTML}{CED7DC}
\definecolor{panelgray}{HTML}{F4F6F7}
\definecolor{trainteal}{HTML}{26756B}
\definecolor{trainlight}{HTML}{EAF5F2}
\definecolor{projectInk}{HTML}{233E53}
\definecolor{projectLine}{HTML}{CCD8E0}
\definecolor{projectTint}{HTML}{F3F7F9}

\tikzset{
  pics/source-code/.style={code={
    \draw (-2,2) -- (-4,0) -- (-2,-2) (2,2) -- (4,0) -- (2,-2)
          (.8,3) -- (-.8,-3);
  }},
  pics/website/.style={code={
    \draw (0,0) circle (3.7) (0,0) ellipse (1.6 and 3.7) (-3.7,0) -- (3.7,0);
  }}
}

\newcommand{\projectBadge}[2]{%
  \begin{tikzpicture}[baseline=(label.base)]
    \node[rounded corners=7pt, draw=projectLine, fill=projectTint,
          text=projectInk, line width=.45pt, inner xsep=7pt, inner ysep=2.5pt,
          font=\normalfont\sffamily\small] (label) {\hspace*{11pt}\strut #2};
    \pic[x=1pt,y=1pt,draw=projectInk,line width=.6pt,line cap=round,line join=round]
      at ([xshift=8.5pt]label.west) {#1};
  \end{tikzpicture}%
}

\hypersetup{colorlinks=true,linkcolor=ink,citecolor=ink,urlcolor=selectblue,
  pdftitle={RACE: Learning Multimodal Search through Referential Abstraction and Compositional Evidence},
  pdfauthor={Eric Tao, Minhao Sun, Liyan Liu, Zhijing Wu, Xiaohan Yu, Lang Mei, Li Huang, Chao Feng, Jinchao Ma, Sichen Kang, Siyu Mo, Chong Chen}}

\newcommand{\imgs}{\mathcal{I}}
\newcommand{\teacher}{\pi_{\mathrm{T}}}
\newcommand{\student}{\pi_{\theta}}

\newcommand{\teachername}{Qwen3.7-Plus}
\newcommand{\studentname}{Qwen3.5-27B}
\newcommand{\ntrain}{12K}
\newcommand{\resultgain}{3.2 points}

\title{\centering RACE: Learning Multimodal Search\\
through Referential Abstraction\\
and Compositional Evidence}
\author{%
  \parbox[t]{\dimexpr\textwidth-2\tabcolsep\relax}{\centering\bfseries
    Eric Tao\textsuperscript{1,*} \quad Minhao Sun\textsuperscript{2,*} \quad
    Liyan Liu\textsuperscript{1,*} \quad Zhijing Wu\textsuperscript{2}\\[0.4ex]
    Xiaohan Yu\textsuperscript{1} \quad Lang Mei\textsuperscript{1} \quad
    Li Huang\textsuperscript{1} \quad Chao Feng\textsuperscript{1}\\[0.4ex]
    Jinchao Ma\textsuperscript{1} \quad Sichen Kang\textsuperscript{1} \quad
    Siyu Mo\textsuperscript{1} \quad Chong Chen\textsuperscript{1,\textdagger}\\[0.8ex]
    \normalfont
    \textsuperscript{1}Huawei \quad \textsuperscript{2}Beijing Institute of Technology\\[0.4ex]
    {\small\textsuperscript{*}Equal contribution. \quad
    \textsuperscript{\textdagger}Corresponding author.}}}
\iclrfinalcopy

\begin{document}
\maketitle
\lhead{Preprint}

\vspace{-0.15in}
\centerline{%
  \href{https://bitminhaosun.github.io/race/}{\projectBadge{website}{project page}}%
  \hspace{5pt}%
  \href{https://github.com/BITMinhaoSun/race}{\projectBadge{source-code}{code}}}
\smallskip

\begin{abstract}
Training multimodal search agents requires tasks that make visually grounded search necessary for determining the answer.
Textual identity clues and unused target evidence allow shortcuts in synthetic questions, weakening supervision for visual search and evidence integration.
We propose \textbf{RACE}, an image-conditioned task synthesis framework that constructs dependencies from images to targets and from target-specific evidence to answers.
\emph{Referential Abstraction} (RA) replaces identity-revealing text with specific visual references or ambiguous relational references resolved by searching among depicted candidates.
\emph{Compositional Evidence} (CE) links retrieved facts about multiple entities in one image or entities from different images into joint questions whose answers require every designated contribution.
Both operations also apply to images found during search, allowing new visual observations to determine subsequent search targets or supply answer evidence.
Fine-tuning \studentname{} on only \ntrain{} teacher-generated trajectories from RACE raises average accuracy across ten multimodal benchmarks from 53.5 to 56.0, a relative improvement of approximately 4.7\% over the matched base agent.
Our code, data, and model will be publicly available.
\end{abstract}

\section{Introduction}
\label{sec:introduction}

Multimodal deep search requires agents to resolve targets in images, search for information about them, and combine evidence.
When search reveals further targets in new images, resolving these references determines what to investigate next.
Recent systems support multi-turn exploration, search guided by intermediate images, and concurrent search across entities \citep{visiondeepresearch,deepvoyagervl,hypereyes}.
Synthetic-data pipelines link evidence across search steps or entities to construct challenging questions, then collect successful teacher trajectories for fine-tuning.
Task structure governs whether answering these questions requires visual target resolution and joint evidence use.

\begin{figure}[t]
  \centering
\includegraphics[width=\textwidth]{\paperdir 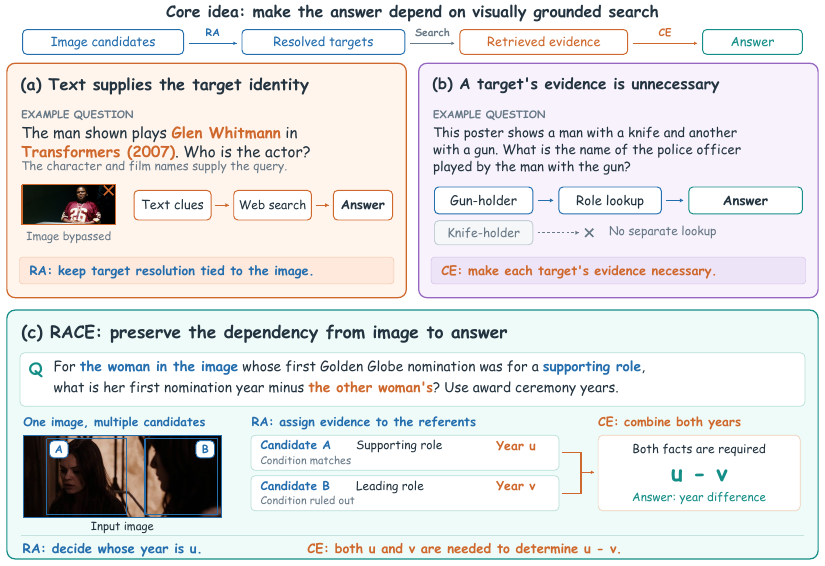}
  \caption{\textbf{Making visually grounded search necessary.}
  (a) Textual identity clues allow the target to be identified without the image.
  (b) The question uses the gun-holder's evidence while leaving the knife-holder's evidence unused.
  (c) RA resolves the relational reference and assigns the two candidates' years to $u$ and $v$; CE combines both years to answer the directed comparison $u-v$.
  Arrows denote information dependencies.}
  \label{fig:abstraction-composition}
\end{figure}

Our central insight is that task synthesis should make visually grounded search necessary for determining the answer.
This requires a dependency from images to targets and from target-specific evidence to the answer.
Textual identity clues let a solver bypass the first dependency by recovering the target without the image (Figure~\ref{fig:abstraction-composition}(a)).
A question can mention two targets while using evidence from just one (Figure~\ref{fig:abstraction-composition}(b)).
These shortcuts bypass visual target resolution or joint evidence use.
In the composed example, resolving the ambiguous reference assigns the two women's facts to a directed comparison, whose answer requires both contributions (Figure~\ref{fig:abstraction-composition}(c)).
In three open-source training sets, 22.3--61.8\% of trajectories both identify the target from the image and obtain evidence for each of multiple relevant entities, compared with 96.2\% for RACE (Appendix~\ref{app:audit}).

We introduce RACE (Referential Abstraction and Compositional Evidence), a framework that constructs these dependencies jointly from real images and retrieved evidence.
\mbox{\emph{Referential Abstraction}} (RA) ties target resolution to the image through specific visual references or ambiguous relational references verified by search.
For ambiguous relational references, construction searches every candidate, supports one match, rules out alternatives, and screens the reference for text-only identity leakage.

\emph{Compositional Evidence} (CE) makes the answer depend on every selected target's evidence.
It connects retrieved facts through supported relations to construct a joint question about multiple entities in one image or one entity from each of several images.
The facts must jointly determine one answer, while removing any designated contribution leaves it undetermined.
RA therefore establishes whose evidence is relevant, and CE specifies how those contributions determine the answer.
This also supports multi-image questions in which every image supplies a necessary contribution.

The two operations recur as text and image searches reveal further targets or evidence, linking visual target resolution to subsequent search and the final answer.
A teacher solves the resulting tasks using only their images and questions.
We retain only trajectories whose final answers are judged correct against the reference answers for supervised fine-tuning (SFT).

We make three contributions:
\begin{itemize}[leftmargin=*,itemsep=3pt,topsep=4pt]
  \item We frame task synthesis around making visually grounded search necessary for the answer. This perspective unifies identity leakage and unused entity evidence as breaks in the dependency from images, through targets and their evidence, to answers.
  \item We introduce RACE, which constructs this dependency through Referential Abstraction and Compositional Evidence: the former ties target resolution to images, while the latter makes every selected target's evidence necessary for joint questions within or across images.
  \item We show that SFT on only \ntrain{} synthesized trajectories raises \studentname{}'s average accuracy from 53.5 to 56.0 across ten benchmarks (Table~\ref{tab:main-results}) and increases visual-tool use beyond the first round on all four primary benchmarks.
\end{itemize}
\section{Related Work}
\label{sec:related-work}

\subsection{Multimodal Deep Search Agents}

Multimodal search agents learn to use image and text tools through supervised training or reinforcement learning (RL) \citep{webwatcher,mmsearchr1,deepmmsearchr1,vsearcher}.
OpenSearch-VL develops fine-grained visual tools \citep{opensearchvl}, Vision-DeepResearch supports multi-turn, multi-entity, and multi-scale retrieval \citep{visiondeepresearch}, and HyperEyes optimizes parallel grounding and search \citep{hypereyes}.
Other methods address visual evidence reuse \citep{lmmsearcher}, context compression by rendering historical text observations as images \citep{pointsseeker}, reasoning over retrieved images \citep{visualseeker}, and adaptive task generation \citep{ode}.
DeepVoyager-VL combines intermediate visual dependencies with selective acquisition of visual evidence for long-horizon search \citep{deepvoyagervl}.
RACE focuses on how task structure determines the reasoning taught by successful trajectories.
It couples visual target discrimination with joint evidence use when constructing tasks for teacher supervision.

\subsection{Datasets and Benchmarks}

Knowledge-based VQA evaluates knowledge beyond image content \citep{okvqa,aokvqa} and fine-grained entity information \citep{infoseek}.
MMSearch and Vision-DeepResearch Benchmark evaluate multimodal retrieval and grounding \citep{mmsearch,visiondeepresearchbench}.
We evaluate on MMSearch-Plus, MM-BrowseComp, BrowseComp-V$^3$, and VisBrowse-Bench, which emphasize provenance, multimodal browsing, and visual evidence verification \citep{mmsearchplus,mmbrowsecomp,browsecompv3,visbrowse}.
To generate training data for these agents, existing synthesis pipelines introduce images into textual tasks or knowledge paths \citep{webwatcher,opensearchvl}, construct questions from complex images \citep{visiondeepresearch}, and incorporate visual-detail steps or webpage images \citep{visualseeker,lmmsearcher}.
Structured approaches use scene graphs \citep{gqa,sage} and image-rooted evidence hypergraphs \citep{mmdeepresearch}.
HyperEyes stitches images from five fine-grained classification datasets into grid mosaics, refers to each tile by its grid position, and requires information from every selected entity \citep{hypereyes}.
DeepVoyager-VL builds event graphs from Wikipedia and news, writes questions with canonical names before controlled masking, and requires minimal constraint sets and retrieved intermediate images whose removal leaves the answer non-unique \citep{deepvoyagervl}.
RACE instead resolves each target among the candidates depicted in a natural scene (Section~\ref{sec:abstraction}).
Its ambiguous relational references select a unique target among several visually plausible candidates through a retrieved relation, and a question-only search probe tests each question for text-only bypass.
RACE formulates synthesis over visually grounded targets, with separate conditions for unique reference resolution and each target's necessary contribution to a joint answer.
Coupling these conditions makes the reasoning requirements explicit at the target level, within scenes, across images, and throughout search.
\section{Method}
\label{sec:method}

\begin{figure}[t]
  \centering
\includegraphics[width=\linewidth]{\paperdir 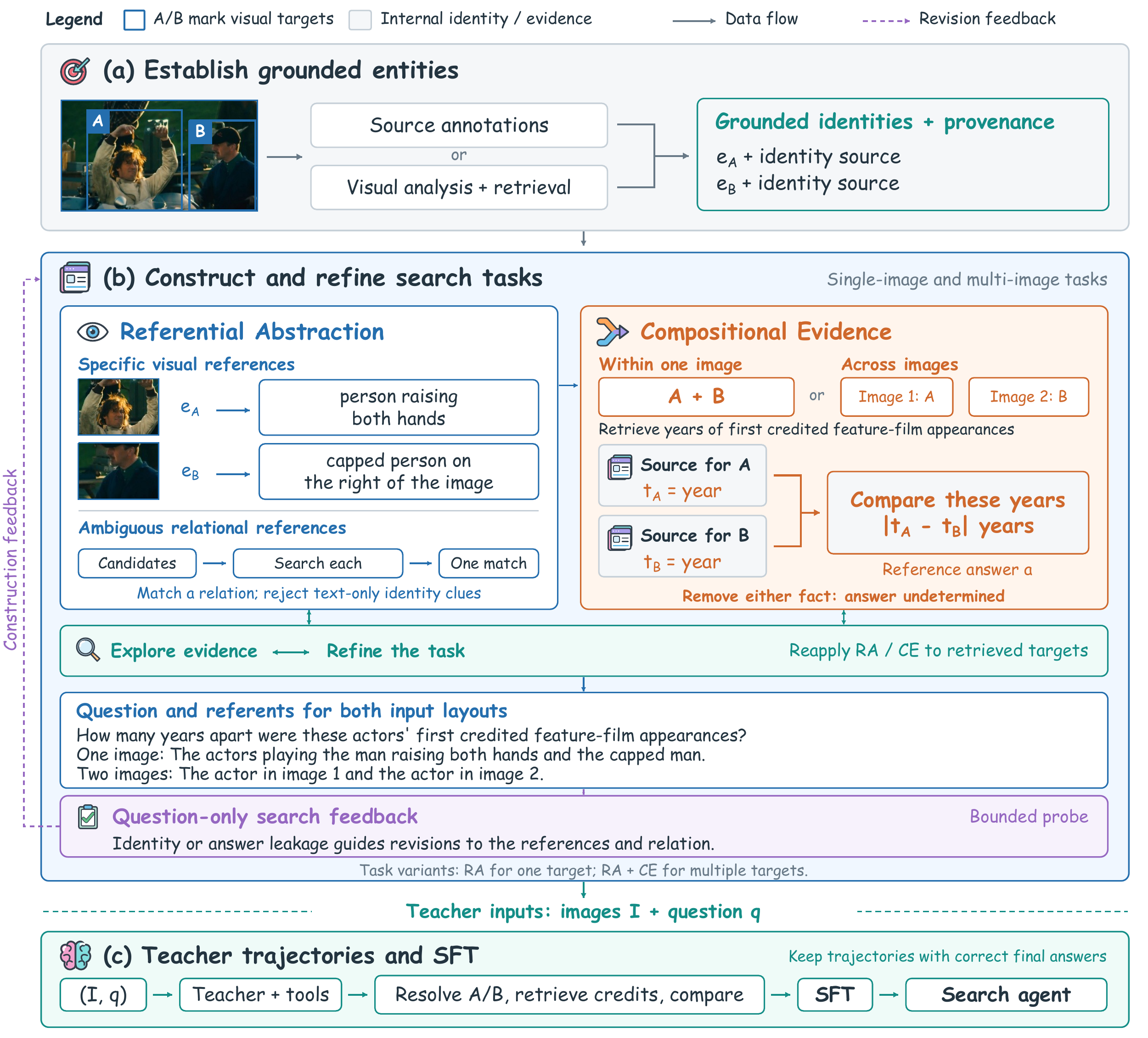}
  \caption{\textbf{RACE task synthesis and training.}
  (a) Source annotations or visual analysis and retrieval establish target identities and provenance.
  (b) RA expresses targets through specific visual or ambiguous relational references. CE combines evidence from multiple targets within or across images. The example compares the years of the actors' first credited feature-film appearances, yielding $|t_A-t_B|$. Evidence exploration and question-only search feedback refine the task.
  (c) The teacher receives $(\imgs,q)$ and uses tools to solve the task; trajectories with correct final answers are retained for SFT.}
  \label{fig:pipeline}
\end{figure}

RACE synthesizes tasks around required dependencies from images to targets and from target-specific evidence to answers (Figure~\ref{fig:pipeline}).
Referential Abstraction ties target resolution to image-grounded references. Compositional Evidence makes each selected target's evidence necessary for the joint answer.
Ambiguous references are resolved through evidence retrieved for the depicted candidates.
Grounding and dependency feedback guide revisions within these operations, including when search introduces new images.
Teacher trajectories with correct final answers provide supervision for SFT.

\subsection{Image-Conditioned Task Synthesis}
\label{sec:formulation}

We start from real images to link target identification in natural scenes to verifiable external evidence.
Sources include social-media photographs, Bilibili video frames, and MovieNet film imagery \citep{huang2020movienet}.
Questions and reference answers are generated from grounded targets and retrieved evidence.
For unannotated images, lightweight filtering, visual analysis, and retrieval establish target identities.
MovieNet annotations provide identity-to-image correspondences.
Both routes supply grounded candidates and identity provenance to the generator.

For input images $\imgs=\{I_1,\ldots,I_m\}$, let $E_i$ denote the candidates in image $I_i$.
The generator selects source targets $S\subseteq\bigcup_i E_i$, constructs references $R=\{\rho_e:e\in S\}$, explores textual and visual evidence $F$, and defines an information requirement $c$.
These inputs determine the generated question and reference answer through
\begin{equation}
  (q,a)=G(\imgs,S,R,c,F), \qquad T=(\imgs,q,a),
  \label{eq:task}
\end{equation}
where $q$ is the question and $a$ its reference answer.
Evidence exploration and question construction can alternate as supported requirements emerge.
Search begins from the source targets and can reveal intermediate targets along the resulting evidence paths.
Solver inputs are $(\imgs,q)$; synthesis records store target identities, reference correspondences, construction evidence, and the reference answer.
A question-only search probe tests whether the question reveals target identities or answers.
The configurations, refinement procedure, and synthesis algorithm are detailed in Appendix~\ref{app:synthesis-details}.

\subsection{Referential Abstraction}
\label{sec:abstraction}

RA supports two forms of reference for candidates $E(V)$ in an initial or retrieved image $V$.
For $\rho=(\nu,\phi)$, $\nu$ specifies visible cues and $\phi$ is an optional condition checked against retrieved candidate evidence $F_e$.
Define the candidate set and resolved matches as
\begin{equation}
  \begin{aligned}
    \mathcal C(V,\nu)&=\{e\in E(V):e\models_V\nu\},\\
    \mathcal M(V,\rho;F)&=\{e\in\mathcal C(V,\nu):\phi(e;F_e)=1\}.
  \end{aligned}
  \label{eq:reference-grounding}
\end{equation}
Both forms require a unique match: $\mathcal M(V,\rho;F)=\{e^\star\}$.

\paragraph{Specific visual references.}
Here $\phi\equiv1$ and $\mathcal C(V,\nu)=\{e^\star\}$: appearance, position, order, spatial relations, or occlusion locate the target before retrieval.
Examples include ``the third person from the left,'' ``the person to the left of the one in blue,'' and ``the person partly hidden behind the post.''

\paragraph{Ambiguous relational references.}
Here $|\mathcal C(V,\nu)|>1$, and $\phi$ selects among visually plausible candidates using retrieved relations or histories.
For example, ``the man in the image who appeared in a crime film in year $y$'' requires checking the candidates' filmographies.
Construction must search every candidate, support one match, and rule out all alternatives; unresolved cases are rejected.
The condition must remain non-unique outside the image; a question-only search probe checks for identity or answer leakage.

RA excludes direct identity disclosures and pairs the resolved reference with a supported information request.
Its candidate searches establish the referent; CE combines facts about resolved targets into a joint answer.

\subsection{Compositional Evidence}
\label{sec:composition}

Compositional Evidence constructs joint questions from retrieved facts about multiple targets.
For source images $I_1,\ldots,I_m$, the selected targets are
\begin{equation}
  S_i\subseteq E_i,\qquad S=\bigcup_{i=1}^{m}S_i,\qquad |S|\geq 2.
  \label{eq:cross-image}
\end{equation}
Within-image composition uses $m=1$ and $|S_1|\geq2$.
Cross-image composition uses $m\geq2$ and $S_i=\{e_i\}$, with a distinct grounded target from each image and $\bigl|\{i:S_i\neq\varnothing\}\bigr|\geq2$.
The generator searches each target and records its facts, sources, and image provenance.
It then connects the evidence through compatible quantities, shared relations, or intersecting constraints.
If no supported joint relation can be constructed, the target or image selection is revised or discarded.

Let $z_1,\ldots,z_k$ denote the selected evidence contributions, with each target supplying at least one.
CE constructs
\begin{equation}
  a=c(z_1,\ldots,z_k), \qquad k\geq 2.
  \label{eq:composition}
\end{equation}
For example, the years of two actors' first credited feature-film appearances support a year-difference question whether they are referenced in one image or separate images.
The generated question refers to every required target and image, and the recorded solution links each contribution to the answer.

Let $[k]=\{1,\ldots,k\}$ and let $\mathcal Y_c(U)$ denote answers consistent with the question and designated facts indexed by $U\subseteq[k]$.
The construction requires joint sufficiency and individual necessity:
\begin{equation}
  \mathcal Y_c([k])=\{a\},\qquad
  \bigl|\mathcal Y_c([k]\setminus\{j\})\bigr|>1
  \quad\text{for every }j\in[k].
  \label{eq:branch-necessity}
\end{equation}
All designated facts determine one answer; removing any one leaves it undetermined under the constructed requirement.
Each image must supply necessary evidence through its target, so removing an image's contributions must also leave the answer undetermined.
CE checks source support and recorded dependencies, revising or rejecting unsupported relations and redundant contributions.
Shortcut probes test additional solution paths under a bounded search budget (Appendix~\ref{app:synthesis-details}).
Both configurations also apply to targets introduced by retrieved images.
We empirically verify these screening and necessity criteria in Section~\ref{sec:trajectory-behavior} and Appendices~\ref{app:probe-stats}, \ref{app:image-removal}, and~\ref{app:audit}.

\subsection{Applying the Operations during Search}
\label{sec:visual-evidence}

A retrieved image can supply answer evidence or determine the next search target.
To obtain answer evidence, the generator retrieves an image $J$ associated with a target and identifies a relevant visual attribute $v(J)$, such as a color, spatial configuration, or inscription.
The question identifies $J$ through its relation to the target and asks the solver to acquire the attribute.
Its value can answer the question directly or contribute to a joint answer through
\begin{equation}
  a=v(J)\qquad\text{or}\qquad a=c\bigl(v(J),z_2,\ldots,z_k\bigr).
  \label{eq:visual-answer}
\end{equation}

When a retrieved scene determines the next search target, the generator applies Referential Abstraction again.
A reference $\rho_W$ selects a target $W$ in an input image.
Retrieval supplies a scene $J$, in which a reference $\rho_{W'}$ selects a further target $W'$ for subsequent search:
\begin{equation}
  W\xrightarrow{\mathrm{retrieve}}J\xrightarrow{\rho_{W'}}W'
    \xrightarrow{\mathrm{search}}z(W').
  \label{eq:visual-transition}
\end{equation}
The retrieved fact $z(W')$ can answer a single-target question or contribute to a composed answer $a=c\bigl(z(W'),z_2,\ldots,z_k\bigr)$.
The reference in $J$ specifies which depicted entity supplies the next information requirement.
The operations also assess whether each required image can be located from preceding evidence and what it establishes for the answer or a later step.
As requirements expand, an earlier answer can become an intermediate fact.
A retrieved image can serve both roles, and information from several newly observed targets can be composed.

\subsection{Teacher Trajectories and Supervised Fine-Tuning}
\label{sec:training}

\teachername{} solves retained tasks with search tools, receiving only their images and questions.
For $N$ retained tasks $T_n=(\imgs_n,q_n,a_n)$, teacher execution and trajectory selection give
\begin{equation}
  \tau_n\sim\teacher(\cdot\mid\imgs_n,q_n),\qquad
  \mathcal D=\{(\imgs_n,q_n,\tau_n):g_{\mathrm{traj}}(\tau_n,a_n)=1\}.
  \label{eq:teacher-trajectories}
\end{equation}
Here $g_{\mathrm{traj}}(\tau_n,a_n)=1$ requires a protocol-valid trajectory whose final answer is judged correct against $a_n$.
Each trajectory records assistant outputs and tool observations in execution order, capturing how visual references guide retrieval and how evidence supports the answer.
We train \studentname{} on \ntrain{} synthesized trajectories in one SFT stage.
For trajectory tokens $u_1,\ldots,u_L$, we minimize the loss over assistant-output positions $\mathcal A(\tau)$,
\begin{equation}
  \mathcal L_{\mathrm{SFT}}(\theta)
  =-\mathbb E_{(\imgs,q,\tau)\sim\mathcal D}
    \left[\sum_{t\in\mathcal A(\tau)}
      \log p_\theta(u_t\mid\imgs,q,u_{<t})\right].
  \label{eq:sft}
\end{equation}
Tool observations provide context, while assistant tool calls and responses provide supervised outputs.
The questions and demonstrations in $\mathcal D$ carry the construction requirements into training.
\section{Experiments}
\label{sec:experiments}

We evaluate RACE under a shared student model and interaction environment.
We first compare answer accuracy before and after SFT, then analyze how training changes tool use and visual revisiting.

\subsection{Experimental Setup}
\label{sec:experimental-setup}

\paragraph{Training and evaluation.}
We train \studentname{} in one SFT stage using only \ntrain{} trajectories generated by \teachername{} from our synthesized tasks.
Assistant tool calls and responses are supervised, while tool observations provide context.
The primary evaluation covers MMSearch-Plus \citep{mmsearchplus}, MM-BrowseComp \citep{mmbrowsecomp}, BrowseComp-V$^3$ \citep{browsecompv3}, and VisBrowse-Bench \citep{visbrowse}.
These benchmarks assess fine-grained visual-cue use, multimodal retrieval and reasoning, and evidence acquisition and verification during browsing.
We report final-answer accuracy under each benchmark's scoring protocol.
Base and RACE results in Table~\ref{tab:main-results} are mean accuracies over four sampled runs per question, with standard deviations given in Appendix~\ref{app:variability}.
Evaluation sets are specified in Appendix~\ref{app:experimental-protocols}.
Supplementary comparisons cover BrowseComp-VL, VDR, MMSearch, SimpleVQA \citep{simplevqa}, LiveVQA \citep{livevqa}, and FVQA \citep{fvqa} (Table~\ref{tab:main-results}).
Avg.\ weights all ten benchmarks equally and is reported only for rows with complete aggregate scores.

\paragraph{Comparisons.}
Baselines include frontier API models, large backbones, and agents trained for multimodal search.
Our Qwen3.5-27B base and SFT agents share test sets, tool access, decoding settings, inference budgets, and scoring settings.
Published agents, including Vision-DeepResearch \citep{visiondeepresearch}, retain their reported model sizes and training regimes.
These agents' scores keep the evaluation protocols of their sources and serve as system-level references (marked in Table~\ref{tab:main-results}), while the final block of Table~\ref{tab:main-results} and Table~\ref{tab:data-comparisons} provide controlled comparisons under identical backbones, tools, and budgets.
Score sources and evaluation variants are recorded in Appendix~\ref{app:experimental-protocols}.

\begin{table}[t]
  \centering
  \caption{\textbf{Search accuracy and factual generalization} (\%, $\uparrow$).
  Results cover four primary search benchmarks and six supplementary benchmarks.
  Models are grouped by role. The final block compares \studentname{} before and after SFT under matched settings.
  Avg. is the unweighted mean over all ten benchmarks.}
  \label{tab:main-results}
  \vspace{4pt}
\begingroup
\fontsize{8}{9.5}\selectfont
\setlength{\tabcolsep}{1.25pt}
\renewcommand{\arraystretch}{1.08}
\begin{tabularx}{\linewidth}{@{}X*{11}{r}@{}}
\toprule[0.7pt]
& \multicolumn{6}{c}{\textcolor{selectblue}{\textbf{Multimodal search}}} & \multicolumn{4}{c}{\textcolor{quietgray}{\textbf{Factual generalization}}} & \\
\cmidrule(lr){2-7}\cmidrule(l){8-11}
\textbf{Model / setting} & \shortstack{MM-\\BC} & MS+ & BC-V$^3$ & VisB & BC-VL & VDR & \shortstack{MM\\Search} & \shortstack{Simple\\VQA} & \shortstack{Live\\VQA} & FVQA & \textbf{Avg.} \\
\midrule
\multicolumn{12}{@{}l}{\textcolor{ink}{\textsf{\textbf{Frontier API and large-backbone references}}}} \\
\addlinespace[2pt]
Qwen3.5-397B-A17B & 46.0 & 42.1 & 31.0 & 52.7 & 49.4 & \textcolor{quietgray}{\textemdash} & 78.3 & 76.4 & 85.2 & 77.1 & \textcolor{quietgray}{\textemdash} \\
Qwen3.7-Plus & 37.1 & 46.3 & 42.3 & 59.8 & 55.6 & 30.2 & 80.7 & 77.7 & 84.6 & 79.7 & 59.4 \\
Qwen3.8-Max & 53.2 & 59.1 & 56.3 & 66.9 & \textcolor{quietgray}{\textemdash} & 33.2 & 84.0 & 83.8 & 88.8 & 85.6 & \textcolor{quietgray}{\textemdash} \\
Gemini-3.1-Pro$^{\ddagger}$ & 48.7 & 51.5 & 50.0 & 62.1 & 65.9 & 40.8 & 86.1 & 81.0 & 87.3 & 84.0 & 65.7 \\
\addlinespace[3pt]
\midrule[0.3pt]
\multicolumn{12}{@{}l}{\textcolor{ink}{\textsf{\textbf{Specialized multimodal search agents (30B--32B)}}}$^{\dagger}$} \\
\addlinespace[2pt]
WebWatcher-32B & \textcolor{quietgray}{\textemdash} & \textcolor{quietgray}{\textemdash} & 8.7$^{\ddagger}$ & \textcolor{quietgray}{\textemdash} & 27.0 & \textcolor{quietgray}{\textemdash} & 55.3 & 59.0 & 58.7 & \textcolor{quietgray}{\textemdash} & \textcolor{quietgray}{\textemdash} \\
SenseNova-MARS-32B & \textcolor{quietgray}{\textemdash} & \textcolor{quietgray}{\textemdash} & \textcolor{quietgray}{\textemdash} & \textcolor{quietgray}{\textemdash} & \textcolor{quietgray}{\textemdash} & \textcolor{quietgray}{\textemdash} & 74.3 & 74.1 & 60.8 & 72.6 & \textcolor{quietgray}{\textemdash} \\
Vision-DeepResearch-30B-A3B & \textcolor{quietgray}{\textemdash} & 28.5 & \textcolor{quietgray}{\textemdash} & \textcolor{quietgray}{\textemdash} & 53.7 & 37.8 & 69.6 & \textcolor{quietgray}{\textemdash} & 77.6 & 74.2 & \textcolor{quietgray}{\textemdash} \\
OpenSearch-VL-30B-A3B & \textcolor{quietgray}{\textemdash} & \textcolor{quietgray}{\textemdash} & \textcolor{quietgray}{\textemdash} & \textcolor{quietgray}{\textemdash} & 41.1 & 33.5 & 68.7 & 74.9 & 67.4 & 73.2 & \textcolor{quietgray}{\textemdash} \\
REDSearcher-MM-RL-30B-A3B & 23.5 & 26.6 & \textcolor{quietgray}{\textemdash} & \textcolor{quietgray}{\textemdash} & 57.2 & \textcolor{quietgray}{\textemdash} & 72.9 & \textcolor{quietgray}{\textemdash} & 79.3 & \textcolor{quietgray}{\textemdash} & \textcolor{quietgray}{\textemdash} \\
Skywork-R1V4-30B-A3B & \textcolor{quietgray}{\textemdash} & \textcolor{quietgray}{\textemdash} & \textcolor{quietgray}{\textemdash} & \textcolor{quietgray}{\textemdash} & 38.4 & \textcolor{quietgray}{\textemdash} & 66.1 & \textcolor{quietgray}{\textemdash} & \textcolor{quietgray}{\textemdash} & 67.2 & \textcolor{quietgray}{\textemdash} \\
MM-DeepResearch-32B & \textcolor{quietgray}{\textemdash} & \textcolor{quietgray}{\textemdash} & \textcolor{quietgray}{\textemdash} & \textcolor{quietgray}{\textemdash} & 43.0 & \textcolor{quietgray}{\textemdash} & 69.0 & 67.6 & 68.0 & 70.1 & \textcolor{quietgray}{\textemdash} \\
LMM-Searcher-30B-A3B (30 turns)$^{\S}$ & 22.3 & 32.9 & \textcolor{quietgray}{\textemdash} & 42.0 & \textcolor{quietgray}{\textemdash} & \textcolor{quietgray}{\textemdash} & 71.0 & \textcolor{quietgray}{\textemdash} & \textcolor{quietgray}{\textemdash} & \textcolor{quietgray}{\textemdash} & \textcolor{quietgray}{\textemdash} \\
\quad same model, 100 turns$^{\S}$ & 30.1 & 34.8 & \textcolor{quietgray}{\textemdash} & 48.3 & \textcolor{quietgray}{\textemdash} & \textcolor{quietgray}{\textemdash} & 72.3 & \textcolor{quietgray}{\textemdash} & \textcolor{quietgray}{\textemdash} & \textcolor{quietgray}{\textemdash} & \textcolor{quietgray}{\textemdash} \\
SimpleSearch-VL-30B-A3B & \textcolor{quietgray}{\textemdash} & 34.4 & \textcolor{quietgray}{\textemdash} & \textcolor{quietgray}{\textemdash} & 55.9 & \textcolor{quietgray}{\textemdash} & 83.6 & 79.6 & 81.1 & 79.0 & \textcolor{quietgray}{\textemdash} \\
\addlinespace[3pt]
\midrule[0.3pt]
\multicolumn{12}{@{}l}{\textcolor{selectblue}{\textsf{\textbf{Our study with the \studentname{} backbone under identical settings}}}} \\
\addlinespace[2pt]
\studentname{} (base agent) & 31.3 & 37.6 & 31.7 & 48.5 & 47.9 & 24.4 & 79.3 & 74.5 & 84.5 & 75.7 & 53.5 \\
\rowcolor{selectlight}
\textbf{RACE-27B (ours)} & \textbf{33.5} & \textbf{40.2} & \textbf{36.7} & \textbf{51.5} & \textbf{54.2} & \textbf{27.4} & 78.7 & \textbf{75.5} & \textbf{85.0} & \textbf{76.8} & \textbf{56.0} \\
\quad $\Delta$  & +2.2 & +2.6 & +5.0 & +3.0 & +6.3 & +3.0 & $-0.6$ & +1.0 & +0.5 & +1.1 & +2.5 \\
\bottomrule[0.7pt]
\end{tabularx}
\par\vspace{4pt}
\begin{minipage}{\linewidth}
\fontsize{8}{9.5}\selectfont
\textit{Notes.} Sources and evaluation settings are detailed in Appendix~\ref{app:experimental-protocols}.
Abbreviations: MM-BC (MM-BrowseComp), MS+ (MMSearch-Plus), BC-V$^3$ (BrowseComp-V$^3$), VisB (VisBrowse-Bench), and BC-VL (BrowseComp-VL); \textemdash{} denotes unavailable scores.
$^{\S}$LMM-Searcher uses Thinking initialization and the MS+ single-image subset, with turn budgets listed by row.
$^{\dagger}$Scores reported in each agent's original paper under its own evaluation setting, shown for reference.
$^{\ddagger}$From the agentic-workflow evaluation compiled by \citet{deepvoyagervl}, except Gemini-3.1-Pro's MM-BC, MMSearch, and FVQA scores, which are our evaluations. $\Delta$ denotes SFT minus base-agent performance, computed from displayed scores, in percentage points.
\end{minipage}
\endgroup
\end{table}

\subsection{Main Results}
\label{sec:main-results}

\paragraph{RACE consistently improves performance over the base agent.}
SFT improves \studentname{} on all four primary benchmarks by 2.2--5.0 percentage points, averaging \resultgain{} under matched evaluation settings (Table~\ref{tab:main-results}).
Accuracy improves on nine of the ten evaluated benchmarks.

\paragraph{Gains are larger on search-oriented benchmarks.}
SimpleVQA, LiveVQA, FVQA, and MMSearch change by $-0.6$ to $+1.1$ points, while the largest improvements occur on the search-oriented BrowseComp-VL and BrowseComp-V$^3$, at 6.3 and 5.0 points, respectively (Table~\ref{tab:main-results}).

\subsection{Training-Data Comparison}
\label{sec:data-configurations}

\paragraph{RACE supervision outperforms the compared open-source training mixture.}
With the same \studentname{} backbone, SFT on \ntrain{} RACE trajectories alone exceeds SFT on the 28K open-source mixture by 0.4--3.0 percentage points across all four primary benchmarks (Table~\ref{tab:data-comparisons}).
The 28K open-source training mixture combines data from OpenSearch-VL, Vision-DeepResearch, and REDSearcher-MM.
An open-source control drawn from this mixture with RACE's size and 41.2\% multi-image share remains below RACE on all four primary benchmarks, by 2.7 points on average (Appendix~\ref{app:visual-structure}).

\paragraph{Adding CE to RA raises the four-benchmark average by 1.3 points.}
With the same teacher, backbone, number of trajectories, and SFT settings, a variant trained only on RA tasks, each with a single target and no compositional evidence, improves the base agent by 1.9 points on average. Adding CE (full RACE) raises the four-benchmark average by a further 1.3 points: by 1.8, 1.4, and 1.8 points on MM-BC, BC-V$^3$, and VisB, and by 0.3 points on MS+ (Table~\ref{tab:data-comparisons}, Appendix~\ref{app:ra-only}).

\subsection{Training-Trajectory Characteristics}
\label{sec:trajectory-behavior}

\paragraph{Training tasks and demonstrations reflect the construction criteria.}
On 499 cross-image composition tasks, a search probe achieves 40.9\% accuracy with both images and 8.2\% on average when one image is withheld (Appendix~\ref{app:image-removal}).
In the teacher-trajectory audit, 96.2\% of RACE trajectories both identify the target from the image and obtain evidence for every relevant entity (Appendix~\ref{app:audit}).

\begin{figure}[!htbp]
  \centering
  \begin{minipage}[t]{0.49\linewidth}
    \vspace{0pt}
    \makeatletter\def\@captype{table}\makeatother
    \caption{\textbf{Training-data comparison} (\%, $\uparrow$).
    Each SFT variant uses only the indicated training data.
    RA only uses the same teacher and number of trajectories as Ours, with single-target tasks (Appendix~\ref{app:ra-only}).}
    \label{tab:data-comparisons}
    \vspace{4pt}
\begingroup
\fontsize{8}{9.5}\selectfont
\setlength{\tabcolsep}{2pt}
\renewcommand{\arraystretch}{1.15}
\begin{tabularx}{\linewidth}{@{}X*{4}{r}@{}}
\toprule[0.7pt]
\textbf{Training setting} & MM-BC & MS+ & BC-V$^3$ & VisB \\
\midrule
\studentname{} (base agent) & 31.3 & 37.6 & 31.7 & 48.5 \\
Open-source only (28K) & 32.1 & 37.9 & 36.3 & 48.5 \\
\midrule[0.3pt]
\leavevmode RA only (\ntrain{}) & 31.7 & 39.9 & 35.3 & 49.7 \\
\textbf{Ours only (\ntrain{})} & \textbf{33.5} & \textbf{40.2} & \textbf{36.7} & \textbf{51.5} \\
\bottomrule[0.7pt]
\end{tabularx}
\endgroup
  \end{minipage}\hfill
  \begin{minipage}[t]{0.49\linewidth}
    \vspace{0pt}
    \centering
    \makeatletter\def\@captype{figure}\makeatother
    \caption{\textbf{Multi-image input.}
    Trajectory proportions across training sources.}
    \label{fig:dataset-analysis}
    \vspace{4pt}
    \includegraphics[width=\linewidth]{\paperdir 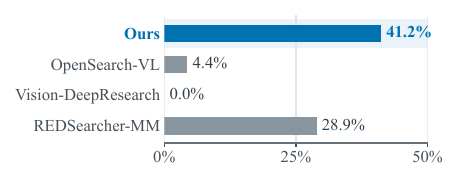}
  \end{minipage}
\end{figure}

\paragraph{RACE trajectories more often use multiple input images.}
At least two images occur in the initial input of 41.2\% of RACE trajectories, compared with at most 28.9\% across the three training-source baselines (Figure~\ref{fig:dataset-analysis}).

\begin{figure}[t]
  \centering
  \includegraphics[width=\linewidth]{\paperdir 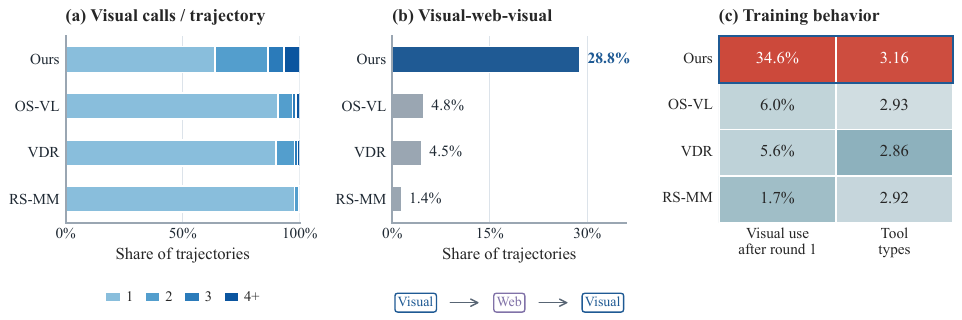}
  \caption{\textbf{Training trajectory distributions and behavior.}
  (a) Visual calls per trajectory. (b) Proportion of trajectories containing a visual-web-visual pattern: a visual call, web search or page fetching, and another visual call in serialized order, with intervening calls allowed.
  (c) The proportion of trajectories with visual calls after the first tool round and mean number of distinct tool types. Colors are standardized within each metric, and the blue outline marks Ours.
  Abbreviations: OS-VL, OpenSearch-VL; VDR, Vision-DeepResearch; RS-MM, REDSearcher-MM.}
  \label{fig:trajectory-distributions}
\end{figure}

\paragraph{Visual calls occur more often after the first round.}
Visual tools comprise image search, text-to-image search, and image cropping, and calls within one assistant message form a tool round.
Visual-tool use extended beyond the first round in 34.6\% of Ours trajectories, compared with 6.0\%, 5.6\%, and 1.7\% for OpenSearch-VL, Vision-DeepResearch, and REDSearcher-MM, respectively (Figure~\ref{fig:trajectory-distributions}(c)).
This higher rate persisted within each of four round-count ranges (Appendix~\ref{app:training-rounds}).

\paragraph{Visual-web-visual sequences are more frequent.}
A visual call followed by web search or page fetching and another visual call occurred in 28.8\% of Ours trajectories, compared with 4.8\%, 4.5\%, and 1.4\% for the same baselines (Figure~\ref{fig:trajectory-distributions}(b)).
This visual-web-visual pattern follows serialized call order and allows intervening calls.
Ours also had a larger share of trajectories with multiple visual calls (Figure~\ref{fig:trajectory-distributions}(a)) and used more distinct tool types per trajectory on average: 3.16 versus 2.86--2.93 (Figure~\ref{fig:trajectory-distributions}(c)).
Metric definitions and supplementary visual-span statistics are provided in Appendices~\ref{app:behavioral-measures} and~\ref{app:training-visual-span}.

\begin{figure}[t]
  \centering
  \includegraphics[width=\linewidth]{\paperdir 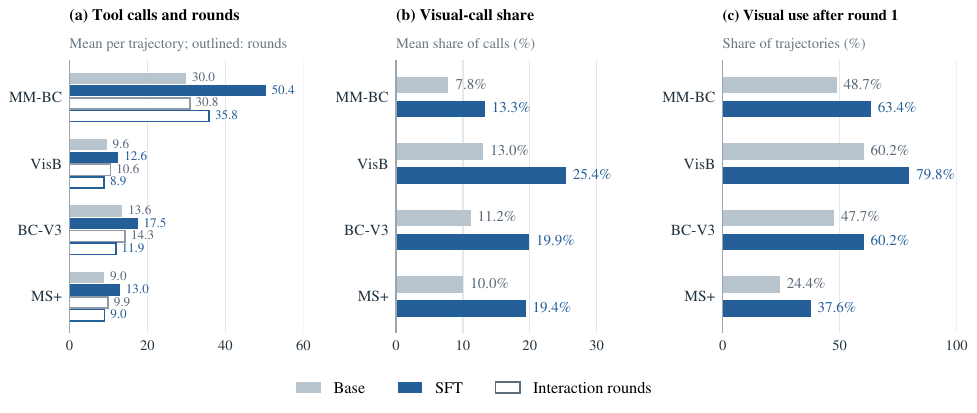}
  \caption{\textbf{Evaluation-time search behavior before and after SFT.}
  Gray and blue bars show base-agent and SFT-agent results, respectively.
  (a) Mean tool calls (filled bars) and interaction rounds (outlined bars) per trajectory. (b) Mean within-trajectory fraction of tool calls that use visual tools.
  (c) Fraction of trajectories that use visual tools after the first tool round.
  Panels (b) and (c) show percentages.
  Abbreviations: MM-BC, MM-BrowseComp; VisB, VisBrowse-Bench; BC-V3, BrowseComp-V$^3$; MS+, MMSearch-Plus.}
  \label{fig:behavior-combined}
\end{figure}

\begin{samepage}
\subsection{Evaluation-Time Search Behavior}
\label{sec:evaluation-behavior}

\paragraph{SFT increases visual-call share and total tool use.}
After SFT, the student used visual tools beyond the first round more often on all four held-out benchmarks, with increases of 12.5--19.6 points (Figure~\ref{fig:behavior-combined}(c)).
Visual-call share, the within-trajectory fraction of visual calls averaged across trajectories, also increased on every benchmark, by 5.5--12.4 points (Figure~\ref{fig:behavior-combined}(b)).
These changes accompanied an increase of 3.0--20.4 tool calls per trajectory, largest on MM-BrowseComp (Figure~\ref{fig:behavior-combined}(a)).

\end{samepage}

\paragraph{Additional calls come mainly from parallel calls within rounds.}
Base and SFT agents share per-benchmark turn limits and may issue multiple tool calls within one round.
On three of the four primary benchmarks, the SFT agent uses fewer interaction rounds than the base agent: 9.0 versus 9.9 on MMSearch-Plus, 11.9 versus 14.3 on BrowseComp-V$^3$, and 8.9 versus 10.6 on VisBrowse-Bench (Figure~\ref{fig:behavior-combined}(a)).
Its additional calls come mainly from parallel calls within a round, at 1.41--1.47 calls per round compared with 0.91--0.97 for the base agent (Appendix~\ref{app:variability}).
At every tool-call budget, RACE achieves higher accuracy than the base agent on MM-BrowseComp and BrowseComp-V$^3$ (Appendix~\ref{app:variability}).

\begin{samepage}
\paragraph{Visual use after the first round increases across task structures.}
Across 790 matched question pairs grouped by dependency depth, checklist length, and input-image count, the SFT agent shows higher post-first-round visual use in all nine strata (Appendix~\ref{app:complexity-strategy}, Tables~\ref{tab:complexity-strategy}--\ref{tab:complexity-strategy-mmsearch} and Figure~\ref{fig:complexity-strategy-delta}).

\end{samepage}

\paragraph{The trained agent relies on returned visual evidence and completes more required steps.}
With the tool set and interaction limits unchanged, masking the returns of image search and text-to-image search lowers RACE's accuracy from 51.5 to 36.1 on VisBrowse-Bench and from 40.2 to 35.2 on MMSearch-Plus.
On MM-BrowseComp, RACE's 2.2-point gain in overall accuracy is matched by a 2.2-point gain in Strict Accuracy, which also requires completing every step of the benchmark's reasoning checklist (Appendix~\ref{app:evidence-use}).

\noindent\begin{minipage}{\linewidth}
\section{Conclusion}
\label{sec:conclusion}

We introduced RACE, an image-conditioned task synthesis framework that constructs dependencies from images to targets and from their evidence to answers.
Referential Abstraction ties target resolution to image-grounded references, and Compositional Evidence connects the resolved targets' evidence to a joint answer that requires every designated contribution.
Both operations apply to initial and retrieved images, so new observations can establish further targets or contribute answer evidence.
Grounding and dependency feedback refine the two operations during synthesis, with question-only probes identifying additional shortcuts.
Fine-tuning \studentname{} on only \ntrain{} teacher-generated trajectories raises average accuracy across ten multimodal benchmarks from 53.5 to 56.0, a gain of 2.5 points over the matched base agent.
\end{minipage}

\subsection*{AI use statement}
Generative AI assisted task synthesis and manuscript preparation, including drafting, language editing, and reference lookup.
In the synthesis pipeline, models ground visual targets, construct questions and reference answers, and support quality checks and shortcut probes (Section~\ref{sec:method} and Appendix~\ref{app:synthesis-details}).
\teachername{} generates the search trajectories used for supervised fine-tuning.
The authors take responsibility for the manuscript, synthesized data, and reported findings.

\subsection*{Ethics statement}
The use and redistribution of RACE's images and web evidence must respect source licenses, platform terms, and the privacy of identifiable people.
Synthetic questions, reference answers, and trajectories carry risks of factual error and bias from source material and generative models.
Construction checks cover task grounding, answer support, and shortcuts.
Applications involving personal information or consequential decisions require privacy and safety review.

\subsection*{Reproducibility statement}
Section~\ref{sec:method} and Appendix~\ref{app:synthesis-details} describe the synthesis procedure, screening, and SFT objective.
Section~\ref{sec:experimental-setup} and Appendix~\ref{app:experimental-protocols} specify the evaluated models, benchmark sets, score sources, decoding settings, and iteration limits.
Appendix~\ref{app:dataset-behavior} documents trajectory selection, behavioral metrics, and uncertainty estimation.
We plan to release the synthesis and analysis code, prompts, and redistributable task and trajectory data, together with the data and scripts underlying the reported tables and figures.
Re-running synthesis and evaluation requires the corresponding models, image assets, and search services; web content, retrieval responses, and hosted models vary over time.

\begingroup
\ExplSyntaxOn
\RenewDocumentCommand{\url}{m}{
  \tl_set:Nn \l_tmpa_tl {#1}
  \regex_replace_once:nnN
    { \A (https://arxiv\.org/abs/[0-9]+\.[0-9]+) v[0-9]+ \Z }
    { \1 } \l_tmpa_tl
  \href{#1}{\exp_args:NV \nolinkurl \l_tmpa_tl}
}
\ExplSyntaxOff
\bibliography{\paperdir references}
\endgroup
\bibliographystyle{\paperdir iclr2027_conference}
\clearpage
\appendix
\raggedbottom
\renewcommand{\topfraction}{0.70}
\renewcommand{\bottomfraction}{0.55}
\renewcommand{\textfraction}{0.25}
\renewcommand{\floatpagefraction}{0.85}
\setlength{\floatsep}{10pt plus 2pt minus 2pt}
\setlength{\textfloatsep}{12pt plus 2pt minus 2pt}
\setlength{\intextsep}{10pt plus 2pt minus 2pt}
\section{Experimental Protocols}
\label{app:experimental-protocols}

The main agent is trained exclusively on \ntrain{} synthesized trajectories.

\subsection{Training and Evaluation Configuration}

\paragraph{Teacher trajectories.}
Each synthesized task contains images, a question, and a supported reference answer.
\teachername{} solves each task with search tools. Its input consists of the images and question; the reference answer and synthesis records support construction and verification.
Only protocol-valid trajectories whose final answers are judged correct against the reference answers are retained.

\paragraph{Supervised fine-tuning.}
We train \studentname{} in one SFT stage on the \ntrain{} retained trajectories.
The objective minimizes negative log-likelihood over assistant tool calls and responses (Equation~\ref{eq:sft}).
The supervised positions are the assistant-output tokens, conditioned on tool observations.

\begin{table}[ht]
  \centering
  \small
  \caption{\textbf{Training setup and evaluation settings.}
  Decoding settings and iteration limits apply to the evaluation runs used for behavioral analysis.}
  \label{tab:implementation-record}
  \begin{tabularx}{\linewidth}{@{}p{0.39\linewidth}X@{}}
    \toprule
    Field & Specification \\
    \midrule
    Teacher and student & \teachername{} and \studentname{} \\
    Main training data & \ntrain{} synthesized trajectories only \\
    Training stage and supervision & One SFT stage over assistant tool calls and responses \\
    Trajectory-run decoding settings & \texttt{temperature = 0.7}, \texttt{top-p = 0.9} \\
    Trajectory-run iteration limits & MM-BC (100), MS+ (20), BC-V$^3$ (20), VisB (15) \\
    Primary benchmark questions & MM-BC (224), MS+ (311), BC-V$^3$ (300), VisB (169), using full sets \\
    \bottomrule
  \end{tabularx}
\end{table}

\paragraph{Benchmark versions and scoring.}
The base and SFT agents use the original 224-question MM-BrowseComp release \citep{mmbrowsecomp} and the full evaluation sets of MMSearch-Plus (311 questions), BrowseComp-V$^3$ (300), and VisBrowse-Bench (169) \citep{mmsearchplus,browsecompv3,visbrowse}.
MMSearch-Plus includes single-image and multi-image questions.
Both agents use the same questions and scoring settings, with final-answer accuracy measured under each benchmark's protocol.
BrowseComp-VL, VDR, MMSearch, SimpleVQA \citep{simplevqa}, LiveVQA \citep{livevqa}, and FVQA \citep{fvqa} provide supplementary comparisons.
Avg.\ in Table~\ref{tab:main-results} is the unweighted mean over all ten benchmarks for rows with complete scores; missing scores are marked with \textemdash{}.
The training-data comparison reports the four primary benchmarks (Table~\ref{tab:data-comparisons}), whose mean gain in Section~\ref{sec:main-results} weights all four equally.

\paragraph{Score sources.}
We report our evaluations of Qwen3.5-397B-A17B, Qwen3.7-Plus, and Qwen3.8-Max.
Our Gemini-3.1-Pro evaluations cover MM-BrowseComp, MMSearch, and FVQA.
WebWatcher, SenseNova-MARS, Vision-DeepResearch, OpenSearch-VL, and MM-DeepResearch use their reported agent scores \citep{webwatcher,sensenovamars,visiondeepresearch,opensearchvl,mmdeepresearch}, as do Skywork-R1V4, REDSearcher-MM-RL, and SimpleSearch-VL \citep{skyworkr1v4,redsearcher,simplesearchvl}.
LMM-Searcher uses Thinking initialization with separate 30- and 100-turn settings. Its MMSearch-Plus scores cover the single-image subset \citep{lmmsearcher}.
Seven Gemini-3.1-Pro scores (MS+, BC-V$^3$, VisB, BC-VL, VDR, SimpleVQA, and LiveVQA) come from the agentic-workflow evaluation of \citet{deepvoyagervl}.
That compilation also supplies WebWatcher's BC-V$^3$ score.

\paragraph{Evaluation runs for behavioral analysis.}
The base and SFT runs analyzed for search behavior share the per-benchmark decoding settings and iteration limits in Table~\ref{tab:implementation-record}.
Both agents use the same system prompt.

\section{Dataset and Behavioral Analysis}
\label{app:dataset-behavior}

\subsection{Metric Computation and Aggregation}
\label{app:behavioral-measures}

\paragraph{Tool normalization.}
Tool aliases are mapped to web search, web fetching and summarization, image search, text-to-image search, and image cropping.
For the measures in Sections~\ref{sec:trajectory-behavior} and~\ref{sec:evaluation-behavior}, calls within an assistant message are counted individually and retain their serialized order.

\paragraph{Per-trajectory statistics.}
Tool-call count is the total number of calls $C$, and visual-call share is the number of visual calls divided by $C$ for $C>0$.
Tool diversity counts the distinct tool types used.
Reported means average the defined per-trajectory values within each population.

\paragraph{Trajectory indicators.}
Post-first-round visual use indicates whether a trajectory contains a visual call after its first tool round.
For the visual-web-visual pattern in Section~\ref{sec:trajectory-behavior}, we test whether serialized call indices $i<j<k$ exist such that calls $i$ and $k$ use visual tools and call $j$ uses web search or page fetching and summarization.
Each trajectory contributes a binary indicator, regardless of how many times the pattern occurs; trajectories without visual calls contribute zero.
For the supplementary visual-search-fetch pattern in Appendix~\ref{app:complexity-strategy}, the three ordered calls correspond to a visual tool, web search, and page fetching.
Both indicators use tool types and serialized call order, allowing intervening calls and calls within one assistant message.
Multi-image input indicates at least two image placeholders in the initial user message.
We report each indicator as the proportion of trajectories satisfying it.

\subsection{Task-Stratified Search Behavior}
\label{app:complexity-strategy}

\paragraph{Population and task strata.}
We match evaluation trajectories by benchmark and query identifier, retaining one saved successful execution per model and question.
Each bin therefore contains the same questions for both models.
Selection uses execution status, independently of answer correctness.
BrowseComp-V$^3$ contributes 284 pairs, grouped by the longest dependency chain in its annotated subgoal graph, counting nodes. The depth bins are 2--3 ($n=107$), 4--5 ($n=123$), and 6 or more ($n=54$).
MM-BrowseComp contributes 195 displayed pairs with 2 ($n=63$), 3 ($n=76$), or at least 4 ($n=56$) checklist items. The seven pairs with one item are omitted.
MMSearch-Plus contributes 311 pairs with 1 ($n=222$), 2 ($n=54$), or at least 3 ($n=35$) input images.
These labels come from benchmark metadata, independently of agent tool use.
VisBrowse-Bench supplies domain labels, so task stratification uses the three benchmarks with structural annotations.

\begin{table}[htb]
  \centering
  \begin{minipage}[t]{0.49\linewidth}
    \vspace{0pt}
    \caption{\textbf{BrowseComp-V$^3$ search behavior} (\%).
    Base versus trained by dependency depth.}
    \label{tab:complexity-strategy}
    \vspace{4pt}
\begingroup
\fontsize{8}{9.5}\selectfont
\setlength{\tabcolsep}{2pt}
\renewcommand{\arraystretch}{1.12}
\begin{tabularx}{\linewidth}{@{}l*{4}{>{\centering\arraybackslash}X}@{}}
\toprule
& \multicolumn{2}{c}{\shortstack{\textbf{Post-first-round}\\\textbf{visual use}}} & \multicolumn{2}{c}{\shortstack{\textbf{Visual-search}\\\textbf{-fetch}}} \\
\cmidrule(lr){2-3}\cmidrule(l){4-5}
\textbf{Dependency depth} & Base & Trained & Base & Trained \\
\midrule
2--3 & 43.9 & 57.0 & 61.7 & 95.3 \\
4--5 & 46.3 & 61.0 & 52.0 & 93.5 \\
6+ & 57.4 & 64.8 & 51.9 & 92.6 \\
\bottomrule
\end{tabularx}
\endgroup
  \end{minipage}\hfill
  \begin{minipage}[t]{0.49\linewidth}
    \vspace{0pt}
    \caption{\textbf{MM-BrowseComp search behavior} (\%).
    Base versus trained by checklist length.}
    \label{tab:complexity-strategy-mmbc}
    \vspace{4pt}
\begingroup
\fontsize{8}{9.5}\selectfont
\setlength{\tabcolsep}{2pt}
\renewcommand{\arraystretch}{1.12}
\begin{tabularx}{\linewidth}{@{}l*{4}{>{\centering\arraybackslash}X}@{}}
\toprule
& \multicolumn{2}{c}{\shortstack{\textbf{Post-first-round}\\\textbf{visual use}}} & \multicolumn{2}{c}{\shortstack{\textbf{Visual-search}\\\textbf{-fetch}}} \\
\cmidrule(lr){2-3}\cmidrule(l){4-5}
\textbf{Checklist items} & Base & Trained & Base & Trained \\
\midrule
2 & 57.1 & 77.8 & 42.9 & 74.6 \\
3 & 43.4 & 55.3 & 51.3 & 71.1 \\
4+ & 46.4 & 62.5 & 53.6 & 75.0 \\
\bottomrule
\end{tabularx}
\endgroup
  \end{minipage}

\end{table}

\begin{table}[htb]
  \centering
  \begin{minipage}[t]{0.49\linewidth}
    \vspace{0pt}
    \caption{\textbf{MMSearch-Plus search behavior} (\%).
    Base versus trained by input-image count.}
    \label{tab:complexity-strategy-mmsearch}
    \vspace{4pt}
\begingroup
\fontsize{8}{9.5}\selectfont
\setlength{\tabcolsep}{2pt}
\renewcommand{\arraystretch}{1.12}
\begin{tabularx}{\linewidth}{@{}l*{4}{>{\centering\arraybackslash}X}@{}}
\toprule
& \multicolumn{2}{c}{\shortstack{\textbf{Post-first-round}\\\textbf{visual use}}} & \multicolumn{2}{c}{\shortstack{\textbf{Visual-search}\\\textbf{-fetch}}} \\
\cmidrule(lr){2-3}\cmidrule(l){4-5}
\textbf{Input images} & Base & Trained & Base & Trained \\
\midrule
1 & 22.5 & 34.2 & 42.3 & 83.3 \\
2 & 27.8 & 51.9 & 44.4 & 98.1 \\
3+ & 31.4 & 37.1 & 34.3 & 88.6 \\
\bottomrule
\end{tabularx}
\endgroup
  \end{minipage}\hfill
  \begin{minipage}[t]{0.47\linewidth}
    \vspace{10pt}
    \raggedright
    \fontsize{8}{9.5}\selectfont
    Tables~\ref{tab:complexity-strategy}--\ref{tab:complexity-strategy-mmsearch} compare matched questions.
    Visual use follows the first tool round. Ordered visual-search-fetch allows intervening calls.
    Sampling and uncertainty estimation are described in Appendix~\ref{app:complexity-strategy}.
  \end{minipage}
\end{table}

\paragraph{Post-first-round visual use.}
Post-first-round visual use increased over the base agent in all nine strata (Tables~\ref{tab:complexity-strategy}--\ref{tab:complexity-strategy-mmsearch} and Figure~\ref{fig:complexity-strategy-delta}), with the largest gain of 24.1 points on two-image MMSearch-Plus questions.

\paragraph{Supplementary call pattern.}
The visual-search-fetch pattern occurred in 92.6--95.3\% of trained-agent trajectories on BrowseComp-V$^3$, compared with 51.9--61.7\% for the base agent (Table~\ref{tab:complexity-strategy}).
Occurrence was also higher in every MM-BrowseComp and MMSearch-Plus stratum, with paired differences of 19.7--31.7 and 41.0--54.3 points, respectively (Tables~\ref{tab:complexity-strategy-mmbc} and~\ref{tab:complexity-strategy-mmsearch}).

\paragraph{Uncertainty estimation.}
For each bin, we compute model-specific proportions and their paired difference.
Percentile bootstrap 95\% intervals use 4,000 resamples.
Difference intervals resample within-question SFT-base differences to preserve pairing.
These intervals quantify sampling variation across questions in the matched evaluation runs.

\begin{figure}[htb]
  \centering
  \includegraphics[width=0.94\linewidth]{\paperdir 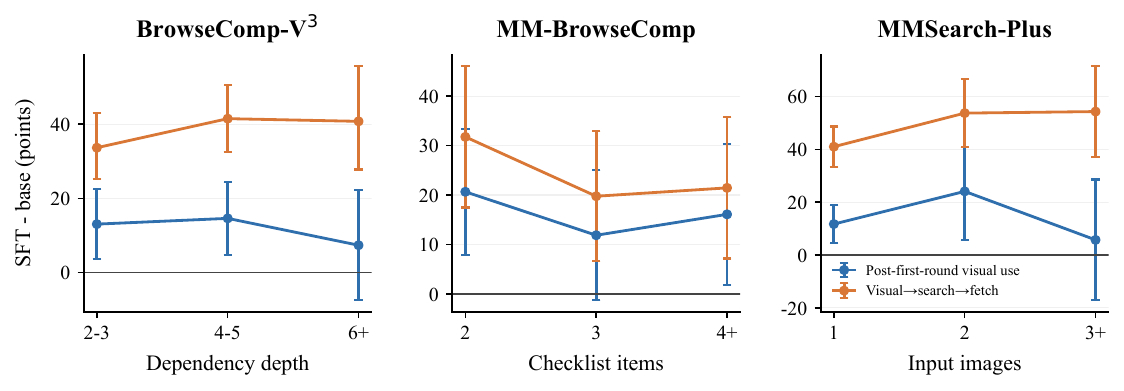}
  \caption{\textbf{Paired differences across task strata.}
  SFT minus base-agent trajectory proportions, in points, for the question pairs in Tables~\ref{tab:complexity-strategy}--\ref{tab:complexity-strategy-mmsearch}.
  Error bars are pointwise paired-bootstrap 95\% intervals. The horizontal line marks zero.}
  \label{fig:complexity-strategy-delta}
\end{figure}

\subsection{Visual Structure of Training Data}
\label{app:visual-structure}

The trajectory analysis compares the analyzed Ours trajectories with 10,752 OpenSearch-VL, 1,393 Vision-DeepResearch, and 16,053 REDSearcher-MM trajectories.
Each trajectory contributes once, with generation and filtering settings retained from its source.

RACE's multi-image share of 41.2\% exceeds the highest baseline share, 28.9\% for REDSearcher-MM (Figure~\ref{fig:dataset-analysis}).
The three baselines contain 470, 0, and 4,643 multi-image trajectories, respectively (5,113 in total).
This count is comparable to the 5,071 multi-image trajectories among the 12,309 RACE training trajectories, whose multi-image inputs come from RACE's cross-image constructions (Appendix~\ref{app:synthesis-details}).
The RA-only variant, whose tasks each have a single target and require no compositional evidence, also exceeds the 28K open-source mixture on average over the four primary benchmarks (39.2 versus 38.7, Table~\ref{tab:data-comparisons}).

\paragraph{Composition-matched open-source control.}
We draw 12,309 trajectories from the 28K open-source mixture with RACE's composition (5,071 multi-image and 7,238 single-image trajectories, a 41.2\% multi-image share), train \studentname{} on them with the same SFT configuration as RACE, and evaluate it with the settings of Table~\ref{tab:implementation-record}.
At the same number of trajectories and the same multi-image share, RACE is 1.8--3.6 points higher on every primary benchmark and 2.7 points higher on average (Table~\ref{tab:composition-control}).
On MMSearch-Plus, where 89 of the 311 questions have multiple input images, the composition-matched control scores the same as the full mixture (37.9).

\begin{table}[ht]

  \centering
  \small
  \caption{\textbf{Composition-matched open-source control} (\%, $\uparrow$). All variants use \studentname{}. The composition-matched control has RACE's number of trajectories and multi-image share.}
  \label{tab:composition-control}
  \setlength{\tabcolsep}{4pt}
  \begin{tabular}{@{}lrrccccc@{}}
    \toprule
    Training data & Traj. & Multi-image & MM-BC & MS+ & BC-V$^3$ & VisB & Avg. \\
    \midrule
    Open-source mixture (full) & 28,198 & 18.1\% & 32.1 & 37.9 & 36.3 & 48.5 & 38.7 \\
    Open-source, composition-matched & 12,309 & 41.2\% & 31.7 & 37.9 & 33.7 & 47.9 & 37.8 \\
    RACE & 12,309 & 41.2\% & \textbf{33.5} & \textbf{40.2} & \textbf{36.7} & \textbf{51.5} & \textbf{40.5} \\
    \bottomrule
  \end{tabular}
\end{table}

\subsection{Continued Visual Use across Interaction Lengths}
\label{app:training-rounds}

We compute continued visual-use rates within four tool-round ranges of 1--5, 6--10, 11--15, and 16 or more.
Rates are reported separately for each source and range.
Ours exceeds the highest baseline rate by 16.8, 34.2, 32.2, and 26.1 points, respectively (Figure~\ref{fig:tool-distributions}).

\begin{figure}[htb]
  \centering
  \includegraphics[width=0.88\linewidth]{\paperdir 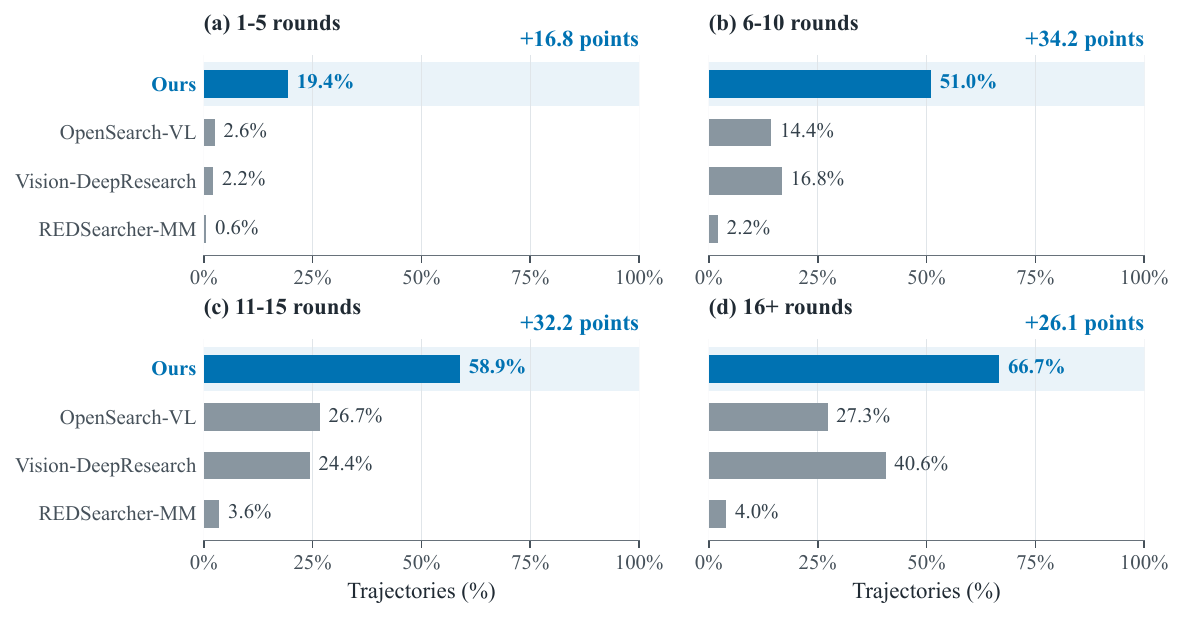}
  \caption{\textbf{Continued visual use across interaction lengths.}
  Bars give the proportion of trajectories with a visual-tool call after the first round, within each round-count range.
  All panels share a scale.
  Blue bars denote Ours, and blue annotations give the difference from the highest baseline rate.}
  \label{fig:tool-distributions}
\end{figure}

\FloatBarrier
\subsection{Supplementary Visual-Span Statistics}
\label{app:training-visual-span}

\begin{figure}[htb]
  \centering
  \includegraphics[width=0.90\linewidth]{\paperdir 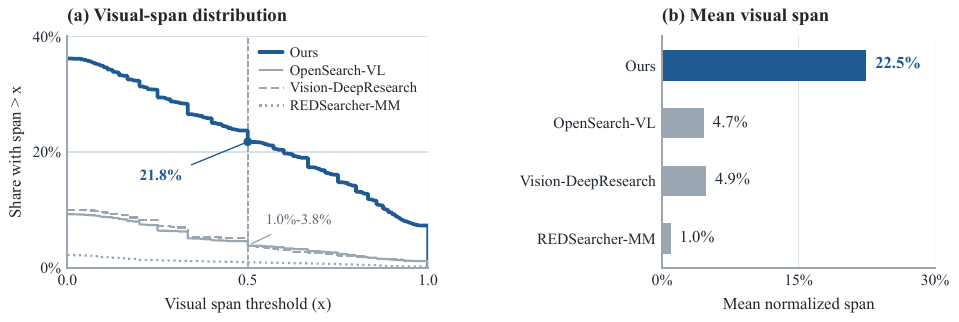}
  \caption{\textbf{Supplementary visual-span statistics for training trajectories.}
  (a) Proportion of trajectories whose normalized visual span exceeds each threshold $x$; higher curves indicate more trajectories with widely separated visual calls.
  (b) Mean normalized visual span for each source.
  Both panels summarize trajectories with at least one visual call; a single visual call has span zero.
  Baseline names follow Table~\ref{tab:main-results}.}
  \label{fig:training-visual-span}
\end{figure}

Normalized visual span measures the separation between the first and last visual calls relative to the entire tool-call sequence.
For a trajectory with $C$ tool calls and first and last visual-call indices $i_{\mathrm{first}}$ and $i_{\mathrm{last}}$, it is
\[
S = \frac{i_{\mathrm{last}}-i_{\mathrm{first}}}{\max(C-1,1)}.
\]
A single visual call gives $S=0$, and visual calls at both ends of a sequence with $C>1$ give $S=1$.
The proportions and means in Figure~\ref{fig:training-visual-span} use trajectories with at least one visual call.

The first and last visual calls spanned more than half the tool-call sequence in 21.8\% of Ours trajectories with visual calls, compared with 1.0--3.8\% for the baselines.
Mean normalized visual span was 22.5\% for Ours, 4.7\% for OpenSearch-VL, 4.9\% for Vision-DeepResearch, and 1.0\% for REDSearcher-MM.

\FloatBarrier
\section{Task Synthesis Details}
\label{app:synthesis-details}

Referential Abstraction and Compositional Evidence support single-target, within-image, and cross-image tasks (Table~\ref{tab:configurations}).
Grounded construction, iterative refinement, and teacher execution form the synthesis and training procedure in Algorithm~\ref{alg:pipeline}.

Source images include Weibo and X/Twitter photographs, Bilibili video frames, and MovieNet film imagery.
Each candidate has a source image and a description locating it in the scene.
Candidate coverage for ambiguous relational references includes every entity matching the coarse visual description.
MovieNet identity annotations can pair frames containing the same person.
Cross-image target selection follows Equation~\ref{eq:cross-image}.
Reference objects provide visual context, while search targets initiate retrieval and intermediate entities can emerge during exploration.
The task specifies required information; the solver chooses the order of identification and retrieval.

\begin{table}[ht]
  \centering
  \small
  \caption{\textbf{Task configurations.}
  Referential Abstraction constructs target references, and Compositional Evidence adds joint information requirements.
  The same configurations apply to targets in retrieved images.}
  \label{tab:configurations}
  \begin{tabularx}{\linewidth}{@{}>{\raggedright\arraybackslash}p{0.21\linewidth}>{\raggedright\arraybackslash}p{0.29\linewidth}>{\raggedright\arraybackslash}X@{}}
    \toprule
    Construction & Target configuration & Required behavior \\
    \midrule
    Single-target\newline abstraction & One target among the candidates in an image & Resolve a specific visual or ambiguous relational reference, then retrieve the requested information. \\
    \addlinespace[5pt]
    Within-image\newline composition & Multiple required targets in one image & Retrieve evidence for each target and derive an answer requiring all designated contributions. \\
    \addlinespace[5pt]
    Cross-image\newline composition & One required target per image across multiple images & Resolve each image's target and combine its evidence with the other images' contributions. \\
    \bottomrule
  \end{tabularx}
\end{table}

\paragraph{Reference construction.}
Specific visual references use appearance, occlusion cues, or positional phrases computed from normalized bounding boxes: a $3\times3$ grid, left-to-right order within a row, or row membership.
Each retained phrase distinguishes its target within the scene; reference neighbors supply spatial context.
For an ambiguous relational reference $\rho=(\nu,\phi)$, the construction record must cover every candidate in $\mathcal C(V,\nu)$ before selecting a unique match.
For each candidate, it must record source evidence and whether the condition is satisfied, ruled out, or unresolved.
Candidates with missing evidence receive the unresolved status.
Retention requires one supported match and supported exclusion of all alternatives; otherwise the generator revises the condition, retrieves further evidence, or discards the task.
Conditions concerning awards, roles, genres, or periods of activity are chosen after examining candidate evidence and are screened for text-only identity leakage.

\paragraph{Target and image constructors.}
When enabled, the within-image multi-anchor constructor selects two grounded targets.
Cross-image construction grounds one target in each of two images and rejects pairs lacking either identity.
The two identities seed separate opening searches. The generator receives both images and the combined evidence, then constructs a relation supported by both targets' facts.
Both images remain attached to the final question, with one image placeholder next to each corresponding reference.
A same-entity route also pairs two views for reference resolution.
Scene records store the selected targets, image associations, spatial cues, and routing decisions.
Evidence records store the query, source URL, title, acquisition turn, and modality.

\paragraph{Evidence and image necessity.}
CE selects the contributions required by the joint question from the retrieved evidence and records their target and image provenance.
For every designated contribution $z_j$, the construction record must explain why the remaining contributions no longer determine a unique answer.
For a required image $I_i$, the same check removes all contributions rooted in its target.
For example, the difference between the years of two actors' first credited feature-film appearances requires both years, whether the actors appear in one image or two.
The generator revises or rejects image pairs with unsupported relations, duplicate evidence, or redundant target contributions.
It stores each necessity judgment and its rationale in the construction record.

\noindent\begin{minipage}{\linewidth}
\begin{algorithm}[H]
  \caption{Image-conditioned task synthesis and agent training}
  \label{alg:pipeline}
  \small
  \begin{algorithmic}[1]
    \Require Image collections, generator $G$, teacher $\teacher$, student $\student$
    \State $\mathcal D\gets\varnothing$
    \For{each source image collection $\imgs$}
      \State Ground candidate entities $E_i$ in each source image $I_i$
      \State Choose $S$ from one or multiple images and ground each selected target
      \State Explore supporting text and visual evidence $F$
      \State Apply RA to construct specific visual or ambiguous relational references $R$
      \State For ambiguous references, search all candidates and verify a unique supported match
      \State For joint tasks, use CE to connect supported facts through a relation $c$
      \State Check that removing any required fact or image leaves the answer undetermined
      \State Apply abstraction and composition as needed to targets in retrieved images
      \State Specify whether each retrieved image supplies answer evidence or a further search clue
      \State Generate or expand $(q,a)$ from $(\imgs,S,R,c,F)$
      \State Probe for solver shortcuts and combine feedback with RA/CE construction criteria
      \If{feedback identifies an unresolved reference, support, dependency, or shortcut issue}
        \State Reapply RA/CE to revise $(R,c,F)$ and reassess $(q,a)$, or discard
      \EndIf
      \If{a valid task is retained}
        \State Execute $(\imgs,q)$ with $\teacher$ to obtain trajectory $\tau$
        \If{$g_{\mathrm{traj}}(\tau,a)=1$}
          \State Add $(\imgs,q,\tau)$ to $\mathcal D$
        \EndIf
      \EndIf
    \EndFor
    \State Train $\student$ on $\mathcal D$ using Equation~\ref{eq:sft}
  \end{algorithmic}
\end{algorithm}

\paragraph{Construction feedback.}
\label{sec:quality}
RA refines references using scene cues and candidate evidence; CE refines relations using source support and the recorded dependencies.
The generator returns a question, answer, visual step, and ordered reasoning steps with dependency links.
For each input image, it also generates a rationale assessing whether the question would still identify the target or determine the answer after removing that image and its referring clause.
Static checks enforce one image placeholder per attached image and reject identity or answer leakage.
A question-only probe generates search queries from $q$ and retrieves web evidence.
A judge compares its output with the reference answer and grounded identities to flag answer recovery, identity recovery, or substantial candidate narrowing.
Detected failures guide revisions to the references, evidence, or relation, followed by renewed checks.
Wording revisions preserve the supported answer; changes to the information requirement trigger new evidence and answer checks.
Let $g_{\mathrm{ref}},g_{\mathrm{sup}},g_{\mathrm{dep}}$ denote verdicts on reference grounding, answer support, and evidence and image necessity, and let $b(T)\in\{0,1\}$ indicate whether the bounded question-only probe detects a shortcut.
The retention criterion is
\begin{equation}
  \mathcal T_{\mathrm{keep}}=
  \bigl\{T:g_{\mathrm{ref}}(T)=g_{\mathrm{sup}}(T)=g_{\mathrm{dep}}(T)=1,\ b(T)=0\bigr\}.
  \label{eq:task-retention}
\end{equation}
\end{minipage}

\subsection{Question-Only Probe Statistics}
\label{app:probe-stats}
The question-only probe runs after RA reference construction and the static identity-leakage checks described under \emph{Construction feedback}.
Because these steps already remove direct identity disclosures, the probe tests whether web search from the question alone can still recover a target identity or the answer.
Of all candidate tasks, 2.4\% were flagged by the probe and retained after revision, and 2.7\% still failed after a second revision and were discarded.

\subsection{Cross-Image Image-Removal Test}
\label{app:image-removal}
We evaluate 499 randomly sampled cross-image composition tasks.
Each task refers to one target in each of two images and requires evidence from both images.
A Qwen3.8-Flash search probe solves each task once with both images, once with image 0 withheld, and once with image 1 withheld, and Qwen3.8-Max judges each final answer against the reference answer.
Withholding either image lowers accuracy from 40.9\% to 8.2\%, averaged over the two removal conditions (Table~\ref{tab:image-removal}).

\begin{table}[ht]

  \centering
  \small
  \caption{\textbf{Cross-image image-removal test} (\%). Accuracy of the search probe on 499 cross-image composition tasks.}
  \label{tab:image-removal}
  \begin{tabular}{@{}lc@{}}
    \toprule
    Input condition & Accuracy \\
    \midrule
    Both images & 40.9 \\
    One image withheld (mean of the two conditions) & 8.2 \\
    \bottomrule
  \end{tabular}
\end{table}

\FloatBarrier

\section{RA-Only Ablation}
\label{app:ra-only}
To separate the contributions of RA and CE, we train a variant on RA tasks only.
Each RA-only task contains a single target and requires no compositional evidence.
The variant uses the same teacher (\teachername{}), backbone (\studentname{}), number of trajectories (\ntrain{}), SFT configuration (Appendix~\ref{app:experimental-protocols}), and evaluation settings as RACE.
RA alone improves the base agent by 1.9 points on average, with gains on all four primary benchmarks.
Adding CE raises the four-benchmark average by a further 1.3 points: by 1.8, 1.4, and 1.8 points on MM-BC, BC-V$^3$, and VisB, and by 0.3 points on MS+ (Table~\ref{tab:ra-only}).

\begin{table}[ht]

  \centering
  \small
  \caption{\textbf{RA-only ablation} (\%, $\uparrow$). All variants use \studentname{}. RA only and RACE use the same teacher and the same number of training trajectories (\ntrain{}).}
  \label{tab:ra-only}
  \begin{tabular}{@{}lcccccc@{}}
    \toprule
    Training data & Trajectories & MM-BC & MS+ & BC-V$^3$ & VisB & Avg. \\
    \midrule
    None (base agent) & \textemdash{} & 31.3 & 37.6 & 31.7 & 48.5 & 37.3 \\
    RA only & \ntrain{} & 31.7 & 39.9 & 35.3 & 49.7 & 39.2 \\
    RACE (RA + CE) & \ntrain{} & \textbf{33.5} & \textbf{40.2} & \textbf{36.7} & \textbf{51.5} & \textbf{40.5} \\
    \bottomrule
  \end{tabular}
\end{table}

\FloatBarrier

\section{Sampling Variability and Interaction Cost}
\label{app:variability}
Base and RACE answer every question four times with the decoding settings and turn limits in Table~\ref{tab:implementation-record}.
Table~\ref{tab:main-results} reports the mean accuracy over the four runs, and Table~\ref{tab:variability} adds the standard deviation across runs.
RACE has higher mean accuracy than the base agent on all four primary benchmarks.

Table~\ref{tab:rounds} reports interaction rounds and tool calls per trajectory for the evaluation runs shown in Figure~\ref{fig:behavior-combined}.
An interaction round is one model generation, and both agents may issue several tool calls within one round.
On MMSearch-Plus, BrowseComp-V$^3$, and VisBrowse-Bench, RACE uses 0.9--2.4 fewer rounds than the base agent and achieves 2.6--5.0 points higher accuracy.
Its additional calls come mainly from parallel calls within a round, consistent with compositional tasks that require retrieving each target separately.

\begin{table}[ht]

  \centering
  \small
  \caption{\textbf{Accuracy over four sampled runs} (\%, mean $\pm$ standard deviation across runs).}
  \label{tab:variability}
  \begin{tabular}{@{}lcc@{}}
    \toprule
    Benchmark & Base & RACE \\
    \midrule
    MM-BC & 31.3 $\pm$ 0.9 & 33.5 $\pm$ 1.1 \\
    MS+ & 37.6 $\pm$ 1.0 & 40.2 $\pm$ 1.4 \\
    BC-V$^3$ & 31.7 $\pm$ 1.6 & 36.7 $\pm$ 1.3 \\
    VisB & 48.5 $\pm$ 1.1 & 51.5 $\pm$ 1.4 \\
    \bottomrule
  \end{tabular}
\end{table}

\begin{table}[ht]

  \centering
  \small
  \caption{\textbf{Interaction rounds and tool calls per trajectory} (Base $\rightarrow$ RACE).}
  \label{tab:rounds}
  \begin{tabular}{@{}lcccc@{}}
    \toprule
    Benchmark & Turn limit & Rounds & Tool calls & Calls per round \\
    \midrule
    MM-BC & 100 & 30.8 $\rightarrow$ 35.8 & 30.0 $\rightarrow$ 50.4 & 0.97 $\rightarrow$ 1.41 \\
    MS+ & 20 & 9.9 $\rightarrow$ 9.0 & 9.0 $\rightarrow$ 13.0 & 0.91 $\rightarrow$ 1.44 \\
    BC-V$^3$ & 20 & 14.3 $\rightarrow$ 11.9 & 13.6 $\rightarrow$ 17.5 & 0.95 $\rightarrow$ 1.47 \\
    VisB & 15 & 10.6 $\rightarrow$ 8.9 & 9.6 $\rightarrow$ 12.6 & 0.91 $\rightarrow$ 1.42 \\
    \bottomrule
  \end{tabular}
\end{table}

\paragraph{Accuracy under tool-call budgets.}
We apply a per-question tool-call budget to the evaluation trajectories behind Table~\ref{tab:main-results}, and a trajectory that has not produced an answer within the budget is counted as incorrect.
At every budget, RACE achieves higher accuracy than the base agent, by 0.8--2.2 points on MM-BrowseComp and 2.7--7.0 points on BrowseComp-V$^3$ (Table~\ref{tab:budget}).

\begin{table}[ht]

  \centering
  \small
  \caption{\textbf{Accuracy under tool-call budgets} (\%, $\uparrow$). A trajectory that has not produced an answer within the budget is counted as incorrect.}
  \label{tab:budget}
  \begin{tabular}{@{}lcccccc@{}}
    \toprule
    MM-BC & $\leq$30 & $\leq$40 & $\leq$50 & $\leq$60 & $\leq$100 & No budget \\
    \midrule
    Base & 29.0 & 31.3 & 31.3 & 31.3 & 31.3 & 31.3 \\
    RACE & 29.9 & 32.1 & 32.1 & 32.6 & 33.5 & 33.5 \\
    RACE $-$ Base & +0.9 & +0.8 & +0.8 & +1.3 & +2.2 & +2.2 \\
    \midrule
    BC-V$^3$ & $\leq$10 & $\leq$13 & $\leq$15 & $\leq$20 & $\leq$30 & No budget \\
    \midrule
    Base & 18.7 & 22.7 & 25.7 & 31.3 & 31.7 & 31.7 \\
    RACE & 23.0 & 29.7 & 30.7 & 34.0 & 36.3 & 36.7 \\
    RACE $-$ Base & +4.3 & +7.0 & +5.0 & +2.7 & +4.6 & +5.0 \\
    \bottomrule
  \end{tabular}
\end{table}

\FloatBarrier

\section{Training-Data Audit}
\label{app:audit}
We randomly sample 500 training trajectories each from OpenSearch-VL, Vision-DeepResearch, REDSearcher-MM, and RACE.
One judge (Qwen3.8-Max) with the same prompt labels every trajectory on two items: (1) whether the target identity is obtained from the image or stated directly in the question text, and (2) for questions involving multiple entities, whether every answer-relevant entity obtains evidence in the trajectory.
Evidence counts only if it comes from tool returns or from reading images. Facts given in the question or assumed by the model do not count.
A trajectory demonstrates both behaviors when the identity is obtained from the image, the question involves multiple entities, and every relevant entity obtains evidence.
All percentages in Table~\ref{tab:audit} use the validly judged trajectories of each source as the denominator.

The differences reflect both task composition and evidence acquisition.
Most OpenSearch-VL and Vision-DeepResearch trajectories involve a single entity, whereas every REDSearcher-MM question involves multiple entities, with only 61.8\% of its trajectories demonstrating both image-based identification and evidence acquisition for every relevant entity.
Among RACE teacher trajectories, 96.2\% demonstrate both behaviors, and 86.8\% do so while verifying at least three entities.

\begin{table}[ht]

  \centering
  \small
  \caption{\textbf{Training-data audit} (\% of validly judged trajectories per source).}
  \label{tab:audit}
  \begin{tabular}{@{}lccc@{}}
    \toprule
    Source & Multi-entity & Both behaviors & Both, $\geq$3 entities verified \\
    \midrule
    OpenSearch-VL & 26.1 & 22.3 & 7.8 \\
    Vision-DeepResearch & 41.6 & 31.8 & 16.2 \\
    REDSearcher-MM & 100.0 & 61.8 & 60.6 \\
    RACE & 99.4 & 96.2 & 86.8 \\
    \bottomrule
  \end{tabular}
\end{table}

\FloatBarrier

\section{Evidence Use of the Trained Agent}
\label{app:evidence-use}
Both analyses apply to the trained agent at evaluation time, and all other settings follow Table~\ref{tab:implementation-record}.

\paragraph{Masking the returns of visual search tools.}
The tool set and interaction limits are unchanged; we mask only the returns of image search, text-to-image search, or both, replacing them with a text placeholder, while image cropping, web search, page fetching, and the input images are unaffected.
Masking both returns lowers RACE's accuracy from 51.5 to 36.1 on VisBrowse-Bench and from 40.2 to 35.2 on MMSearch-Plus, and masking either one alone also lowers it on both benchmarks (Table~\ref{tab:blank-visual}).
These results show that the trained agent relies on the information returned by its visual search tools.

\paragraph{Judging the benchmarks' annotated required steps.}
MM-BrowseComp provides a verified irreducible reasoning checklist for every question, listing the steps required to reach the answer; its official Strict Accuracy counts a question as correct only when the answer is correct and every checklist step is completed.
BrowseComp-V$^3$ provides expert-validated subgoals for every question.
Using both benchmarks' official judging protocols and the same judge for both agents, we judge every step on the evaluation trajectories behind Table~\ref{tab:main-results}.
RACE improves Strict Accuracy on MM-BrowseComp by 2.2 points (21.9\% $\rightarrow$ 24.1\%), matching its overall accuracy gain, while the percentage of questions answered correctly with incomplete checklists remains at 9.4\% (Table~\ref{tab:checklist}).
RACE also completes more checklist steps overall (53.0 $\rightarrow$ 55.3), and on BrowseComp-V$^3$ it completes more of the expert-validated subgoals (56.0 $\rightarrow$ 62.5).

\begin{table}[ht]

  \centering
  \small
  \caption{\textbf{Masking the returns of visual search tools} (RACE, \%). The tool set and interaction limits are unchanged; image cropping, web search, page fetching, and the input images are unaffected in every setting.}
  \label{tab:blank-visual}
  \begin{tabular}{@{}lcc@{}}
    \toprule
    Masked returns & VisB & MS+ \\
    \midrule
    None & 51.5 & 40.2 \\
    Image search & 43.8 & 35.7 \\
    Text-to-image search & 36.7 & 36.1 \\
    Both & 36.1 & 35.2 \\
    \bottomrule
  \end{tabular}
\end{table}

\begin{table}[ht]

  \centering
  \small
  \caption{\textbf{Required steps on the evaluation trajectories} (\%).}
  \label{tab:checklist}
  \begin{tabular}{@{}lccc@{}}
    \toprule
    Metric & Base & RACE & Change \\
    \midrule
    MM-BC accuracy & 31.3 & 33.5 & +2.2 \\
    \quad of which: every checklist step completed (Strict Accuracy) & 21.9 & 24.1 & +2.2 \\
    \quad of which: correct with some checklist steps missing & 9.4 & 9.4 & 0.0 \\
    MM-BC checklist score & 53.0 & 55.3 & +2.3 \\
    BC-V$^3$ accuracy & 31.7 & 36.7 & +5.0 \\
    BC-V$^3$ subgoal completion & 56.0 & 62.5 & +6.5 \\
    \bottomrule
  \end{tabular}
\end{table}
\end{document}